\documentclass[conference]{IEEEtran}
\IEEEoverridecommandlockouts
\usepackage{cite}
\usepackage{amsmath,amssymb,amsfonts}
\usepackage{algorithmic}
\usepackage{graphicx}
\usepackage{textcomp}
\usepackage{xcolor}
\usepackage{booktabs}
\usepackage{multirow}
\usepackage{algorithm}
\usepackage{subcaption}

\makeatletter
\renewcommand{\footnoterule}{%
  \kern-3pt
  \hrule\@width 1in 
  \@height 0.4pt 
  \kern 2.6pt
}

\makeatletter
\newcommand{\linebreakand}{%
  \end{@IEEEauthorhalign}
  \hfill\mbox{}\par
  \mbox{}\hfill\begin{@IEEEauthorhalign}
}
\makeatother

\def\BibTeX{{\rm B\kern-.05em{\sc i\kern-.025em b}\kern-.08em
    T\kern-.1667em\lower.7ex\hbox{E}\kern-.125emX}}
\begin{document}
\title{Corrected with the Latest Version: Make Robust Asynchronous Federated Learning Possible\\
\thanks{*Corresponding author: chaoyi@stu.xjtu.edu.cn}
}

\author{
\IEEEauthorblockN{1\textsuperscript{st} Chaoyi Lu*}
\IEEEauthorblockA{\textit{School of Software Engineering} \\
\textit{Xi'an Jiaotong University}\\
Xi'an, China \\
chaoyi@stu.xjtu.edu.cn}
\and
\IEEEauthorblockN{2\textsuperscript{nd} Yiding Sun}
\IEEEauthorblockA{\textit{School of Software Engineering} \\
\textit{Xi'an Jiaotong University}\\
Xi'an, China \\
sunyiding@stu.xjtu.edu.cn}
\and
\IEEEauthorblockN{3\textsuperscript{rd} Pengbo Li}
\IEEEauthorblockA{\textit{International School} \\
\textit{Beijing University of Posts and Telecommunications}\\
Beijing, China \\
2023213418@bupt.cn}
\linebreakand
\IEEEauthorblockN{4\textsuperscript{th} Zhichuan Yang}
\IEEEauthorblockA{\textit{School of Software Engineering} \\
\textit{Xi'an Jiaotong University}\\
Xi'an, China \\
zhichuan@stu.xjtu.edu.cn}
}

\maketitle
\begin{abstract}
As an emerging paradigm of federated learning, asynchronous federated learning offers significant speed advantages over traditional synchronous federated learning. Unlike synchronous federated learning, which requires waiting for all clients to complete updates before aggregation, asynchronous federated learning aggregates the models that have arrived in real-time, greatly improving training speed. However, this mechanism also introduces the issue of client model version inconsistency. When the differences between models of different versions during aggregation become too large, it may lead to conflicts, thereby reducing the model's accuracy. To address this issue, this paper proposes an asynchronous federated learning version correction algorithm based on knowledge distillation, named FedADT. FedADT applies knowledge distillation before aggregating gradients, using the latest global model to correct outdated information, thus effectively reducing the negative impact of outdated gradients on the training process. Additionally, FedADT introduces an adaptive weighting function that adjusts the knowledge distillation weight according to different stages of training, helps mitigate the misleading effects caused by the poorer performance of the global model in the early stages of training. This method significantly improves the overall performance of asynchronous federated learning without adding excessive computational overhead. We conducted experimental comparisons with several classical algorithms, and the results demonstrate that FedADT achieves significant improvements over other asynchronous methods and outperforms all methods in terms of convergence speed.
\end{abstract}


\section{Introduction}
Federated learning \cite{kairouz2021advancesopenproblemsfederated,Liu_2022}, as a machine learning paradigm that combines privacy protection, is increasingly gaining widespread attention. It effectively addresses the issue of data privacy leakage by replacing data transmission with model transmission. Compared to traditional local training, it can significantly improve model accuracy in certain cases. The working principle of federated learning is that a central server coordinates clients to participate in training. In each round of training, the server selects a subset of clients and broadcasts the global model to them. Then, the clients use their local data for training. Once the clients finish training, the server aggregates the model parameters uploaded by all the clients to generate an updated global model.

However, the traditional synchronous federated learning method has a significant drawback: Each round of training must wait for all selected clients to complete their training, which can lead to severe delays, especially when there is a large performance gap between clients. Faster clients often have to wait for slower clients to finish training, a situation known as the \textit{``straggler problem''}. The root cause of the straggler problem is system and communication heterogeneity~\cite{9093123,10.5555/3618408.3618577}, specifically the differences in computing power and bandwidth among clients~\cite{8270639}, which lead to performance imbalances and exacerbate waiting times.

To address the straggler problem, asynchronous federated learning~\cite{XU2023100595,nguyen2021federatedlearningsmarthealthcare} has been introduced. Unlike synchronous methods, asynchronous federated learning does not wait for all clients to complete training. Instead, it aggregates the gradients from any client as soon as they upload them. However, while asynchronous methods effectively mitigate the straggler problem, the lack of synchronization often affects the accuracy of the model. In asynchronous federated learning, since clients may be at different stages of training, the server may aggregate models from multiple versions during each update. This asynchronous updating process can introduce outdated information, which in turn impacts the training effectiveness~\cite{9812885}.

We observe that the primary difference between asynchronous and synchronous methods lies in how updates are made. In synchronous federated learning, each update is based on the most recent model. The server patiently waits for all selected clients to complete training, and uses the latest global model to guide the server in generating high-value gradients with its own training data. In contrast, asynchronous federated learning lacks such patient guidance: \textbf{The server performs multiple updates while the selected clients are still training, resulting in almost abandoned waiting for the clients. Ultimately, the server hastily merges the outdated results after the client has completed training}. We believe this strategy is unreasonable. Although asynchronous federated learning, by its very nature, cannot wait for clients to finish training, it still requires a ``guide" to direct outdated information toward the latest version, thus reducing the impact of outdated information on accuracy.

In this paper, we propose a version correction algorithm for asynchronous federated learning, which utilizes the latest global model to guide and correct outdated information, aiming to improve both the accuracy and performance of asynchronous federated learning. The main contributions of this paper are as follows:

\begin{itemize}
    
\item We propose FedADT, an asynchronous federated learning gradient correction algorithm based on knowledge distillation. After the client sends the training results, the server corrects outdated information using the globally updated model, thereby significantly improving the performance of asynchronous federated learning.

\item We design an adaptive weight-based strategy for FedADT, enabling the client to be better guided by the globally updated model at different stages of training, thus avoiding the potential misguidance of the student model by the teacher model in the early stages of training.

\item We conducted experimental comparisons with several existing methods on computer vision tasks. The experimental results demonstrate that the proposed method exhibits competitive performance across various evaluation metrics and shows significant advantages over other asynchronous methods.
\end{itemize}

\section{Related Work}
\subsection{Synchronous Federated Learning}
Synchronous methods are the leading strategy in federated learning, where the server selects a subset of clients to participate in training during each communication round. The server aggregates the uploaded model data only after all selected clients have completed their training. These synchronous approaches offer notable advantages in terms of model accuracy. Among them, FedAvg \cite{mcmahan2023communicationefficientlearningdeepnetworks} is the most traditional algorithm in synchronous federated learning, creating a new global model each communication round by averaging the models from all participating clients. In contrast, FedAdam \cite{reddi2021adaptivefederatedoptimization} enhances FedAvg by incorporating a momentum mechanism to further improve optimization performance.

Another key benefit of synchronous federated learning is that clients consistently train using the latest global model, ensuring that the server receives the most current gradient information. As a result, synchronous algorithms can efficiently gather the latest client data and optimize the global model to address various complex challenges. For example, FedProx \cite{li2020federatedoptimizationheterogeneousnetworks} introduces a regularization term during local client training to prevent model parameters from deviating too much from the global model. FedBN \cite{li2021fedbnfederatedlearningnoniid} reduces feature shifts caused by differences in data distributions through Batch Normalization. Additionally, MOON \cite{li2021modelcontrastivefederatedlearning}  utilizes the similarity between model representations to adjust the local training processes of individual clients.
\subsection{Asynchronous Federated Learning}
Asynchronous Federated Learning was designed to overcome the issue of ``stragglers'', where the performance of faster clients is hindered by slower ones, thereby limiting overall efficiency \cite{xu2023asynchronousfederatedlearningheterogeneous}. In an entirely asynchronous framework, the global model is updated as soon as any client submits an update, meaning clients don’t have to wait for others. This feature allows asynchronous methods to significantly accelerate training compared to synchronous ones, especially in scenarios with varying client latencies. However, one major drawback of asynchronous systems is the problem of stale updates, which can negatively impact both model accuracy and stability, particularly when working with non-i.i.d. data~\cite{9562538}. To address this challenge, several solutions have been proposed. For example, FedAsync \cite{xie2020asynchronousfederatedoptimization} adjusts the contribution of updates based on their age, FedASMU \cite{liu2023fedasmuefficientasynchronousfederated} ensures clients always use the most current global model during training, and 
FedFa \cite{10.24963/ijcai.2024/584} utilizes a queue-based buffer that discards the oldest update when a new one arrives after the buffer is full.

Additionally, some research has proposed the use of buffers at the server to collect and aggregate client updates, a technique known as semi-asynchronous federated learning. In semi-asynchronous federated learning, gradient aggregation occurs when the buffer reaches its capacity. This approach reduces the frequency of global model updates, thereby mitigating the negative impact of stale updates. For instance, FedBuff~\cite{nguyen2022federatedlearningbufferedasynchronous} empties the buffer once it reaches capacity, while CA2FL~\cite{wang2024tackling} stores the latest updates from each client to better refine the global model. The buffer strategy in semi-asynchronous federated learning performs relatively well in high-latency environments. However, the introduction of buffers increases time consumption, making this solution still unsatisfactory \cite{9093123,10097251}. We aim to design a purely asynchronous approach that can achieve the same or even slightly higher accuracy compared to semi-asynchronous methods, while ensuring training speed, thus achieving a more robust asynchronous federated learning solution.
\subsection{Knowledge Distillation}
The concept of knowledge distillation was introduced by Hinton et al. \cite{hinton2015distillingknowledgeneuralnetwork}, where a large and complex network, referred to as the teacher, is used to generate soft labels. These labels then guide the training of a smaller, more efficient student network. While early approaches focused on transferring knowledge from a single teacher model, more recent techniques have explored the use of multiple teachers and various aggregation methods. These include gate learning strategies in supervised environments \cite{asif2020ensembleknowledgedistillationlearning, xiang2020learningmultipleexpertsselfpaced} and methods that leverage sample similarity in unsupervised settings. Advances in data-free knowledge distillation \cite{zhu2021datafreeknowledgedistillationheterogeneous} emphasize adversarial techniques that generate challenging samples that are difficult for both the student and the teacher to learn from. In a similar vein, DeepInversion \cite{yin2020dreamingdistilldatafreeknowledge} exploits backpropagated gradients to create adversarial samples that induce disagreements between the teacher and the student. Furthermore, \cite{nayak2019zeroshotknowledgedistillationdeep} approaches the task by fitting data distributions based on output similarities to craft a suitable transfer set.

In the context of federated learning based on distillation, there are approaches beyond the traditional parameter-based FL methods \cite{mcmahan2023communicationefficientlearningdeepnetworks, li2020sampleknowledgedistillationefficient}. Early efforts, such as \cite{jeong2023communicationefficientondevicemachinelearning}, involved exchanges of both parameters and model outputs. More recent approaches \cite{li2021practicaloneshotfederatedlearning, li2019fedmdheterogenousfederatedlearning, chang2019cronusrobustheterogeneouscollaborative} focus solely on transferring knowledge via local model outputs. However, these methods typically assume that the data distributions are similar across different devices, creating a reliance on prior knowledge of private data. Some newer methods \cite{NEURIPS2020_18df51b9, Gong_Sharma_Karanam_Wu_Chen_Doermann_Innanje_2022} relax this requirement to some extent, but they still necessitate careful selection of transfer data based on local task specifics and private data characteristics.

Although approaches like \cite{pmlr-v139-zhu21b, 9912292, zhang2023finetuningglobalmodeldatafree} allow knowledge transfer without requiring actual data, they often entail significant communication overhead due to the need for extensive model exchanges over multiple iterations. This makes them vulnerable to potential security threats and privacy risks.

\section{Problem Formulation}

This section explains the problem formulation and system model for federated learning. We consider a system comprising a central server and \( M \) heterogeneous devices (also referred to as edge devices), where each device has its own local private dataset. The goal is to collaboratively train a global model shared by all devices.

\subsection{Problem Definition}

Assume there are  $M$ devices in the system, and the set of devices is denoted by $ \mathcal{M} = \{1, 2, \dots, M\} $. For any device $ i \in \mathcal{M} $, its local dataset can be expressed as:
\begin{equation}
D_i = \{(x_{i,d}, y_{i,d}) \mid d = 1, 2, \dots, |D_i|\},
\end{equation}
where \( x_{i,d} \in \mathbb{R}^s \) represents the $d$-th input sample, \( y_{i,d} \in \mathbb{R} \) represents the corresponding label, and $ |D_i| $ is the number of samples on device $ i $. After aggregating the datasets of all devices in the system, we obtain the global dataset:
\begin{equation}
D = \bigcup_{i \in \mathcal{M}} D_i, \quad N = \sum_{i \in \mathcal{M}} |D_i|,
\end{equation}
where $N$ represents the total number of samples across all devices.

Our goal is to collaboratively train a global model using the local datasets of all devices, without directly exchanging the original data. We define the global loss function as:
\begin{equation}
J(w) = \frac{1}{N} \sum_{i \in \mathcal{M}} \sum_{(x_{i,d}, y_{i,d}) \in D_i} \ell(w; x_{i,d}, y_{i,d}),
\end{equation}
where $ \ell(\cdot) $ represents the loss for a single sample. The local loss function on device $ i $ is defined as:
\begin{equation}
J_i(w) = \frac{1}{|D_i|} \sum_{(x_{i,d}, y_{i,d}) \in D_i} \ell(w; x_{i,d}, y_{i,d}).
\end{equation}

Finally, the Federated Learning problem can be expressed as the following optimization problem:
\begin{equation}
\min_w J(w) = \min_w \frac{1}{N} \sum_{i \in \mathcal{M}} |D_i| J_i(w).
\end{equation}

Synchronous federated learning requires waiting for all selected clients to complete their local training before the server performs model aggregation. In contrast, asynchronous federated learning overcomes this limitation by immediately aggregating the model whenever the server receives an update from any client, thereby generating an updated global model. The aggregation process in asynchronous federated learning can be expressed by the following formula:

\begin{equation}
\mathbf{w}_i^{(g)} = \beta \mathbf{w}^{(l)} + (1 - \beta) \mathbf{w}_{i-1}^{(g)},
\end{equation}
here, $\mathbf{w}^{(l)}$ represents the received client model, $\mathbf{w}_i^{(g)}$ denotes the $i$-th version of the global model, and $\beta$ depends on the value of the received model. Typically, a smaller $\beta$ is assigned to gradients with significant staleness.

\begin{figure}[t]
    \hspace{-0.3cm}
    \centering
    \includegraphics[width=0.5\textwidth]{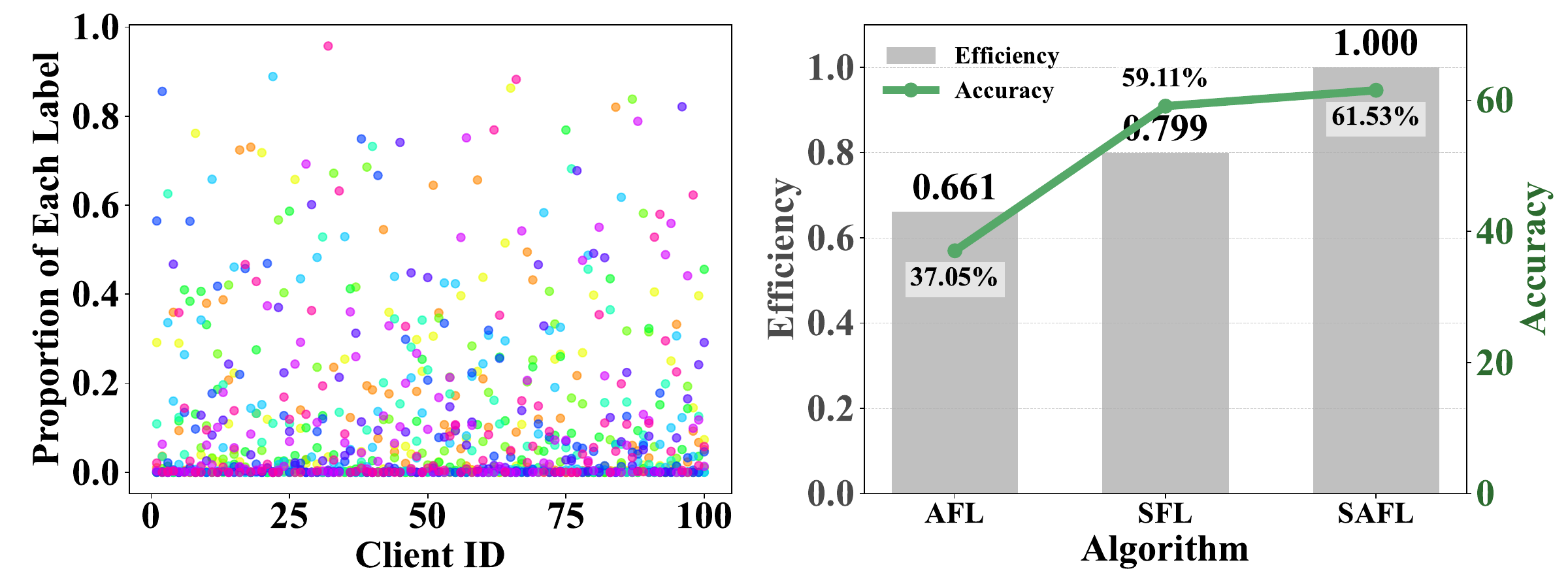} 
    \caption{\textbf{The comparison of the experimental data distribution and algorithm performance.} In the left figure, different colors of points represent different categories, and the position of each point reflects the proportion of each category's data in the corresponding client. In the figure on the right, `Efficiency' refers to the ratio of the time taken by the optimal algorithm to reach a target accuracy of 35\% to the time taken by each algorithm to achieve the same level of accuracy.}
    \label{fig1}
\end{figure}

\begin{algorithm}[t]
\caption{Client-Training}
\textbf{Input:} server model $w$, client learning rate $\eta$, client steps $Q$, global timestamp $t$\\
\textbf{Output:} client model $w_Q$
\begin{algorithmic}[1]
\STATE $y_0 \gets w$
\FOR{$q = 1$ to $Q$}
    \STATE $y_q \gets y_{q-1} - \eta g_q(y_{q-1})$
\ENDFOR
\STATE $w_Q \gets y_Q$ 
\STATE \textbf{Send} $(w_Q,t)$ to server
\end{algorithmic}
\end{algorithm}

\begin{figure*}[t]
    \centering
    \includegraphics[width=0.88\textwidth]{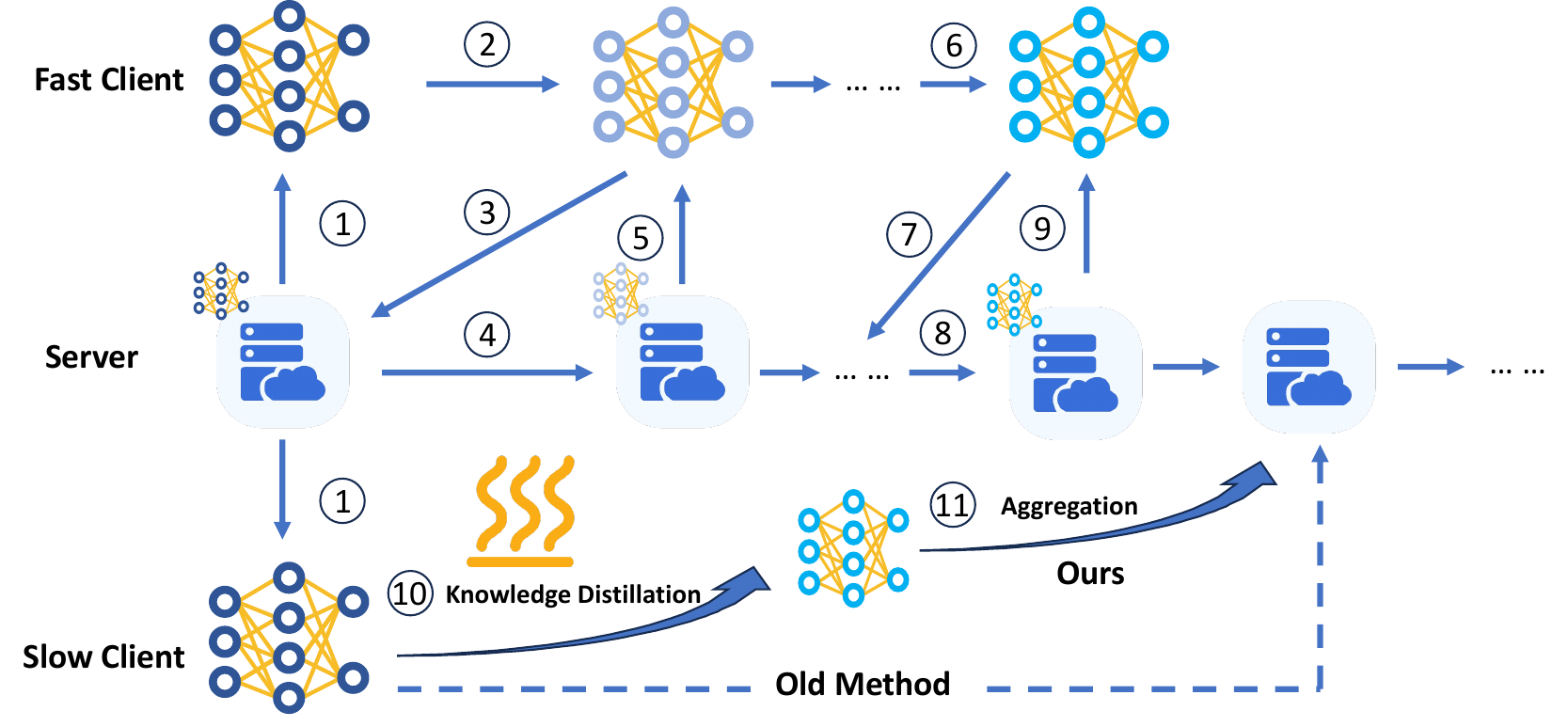}  
    \caption{\textbf{FedADT Framework.} Faster client models are almost always trained based on the latest global model, whereas slower client models can only be trained on outdated global models. Traditional methods typically aggregate these outdated models directly with the latest model, which can lead to conflicts between different model versions. In contrast, our approach leverages knowledge distillation to rapidly align outdated client models with the latest version, significantly reducing conflicts between different model versions.}  
    \label{fig:figure1}
\end{figure*}

\subsection{Dilemma of Asynchronous Federated Learning}
Although asynchronous federated learning has the advantage of faster training speed, it has to simultaneously deal with staleness and the heterogeneity of data distribution, which results in poor performance when handling uneven data distributions. As a result, its training outcomes are often significantly worse than those of other federated learning algorithms. To simulate strong data distribution heterogeneity, we employed a Dirichlet distribution with $\alpha$ = 0.2 to evenly distribute the CIFAR-10 dataset across 100 clients. Based on this, we selected two algorithms for each of the synchronous, semi-asynchronous, and asynchronous federated learning approaches and calculated their average performance, as shown in Fig.~\ref{fig1}. In the case of uneven data distribution, asynchronous federated learning consistently performs worse in terms of accuracy and convergence compared to the other algorithms, leading to its frequent neglect.

\section{Method}
FedADT consists of two main components: The first component is knowledge distillation, which is based on the idea of using the latest version of the global model to correct outdated models uploaded by clients. By introducing this knowledge distillation mechanism, FedADT ensures that the client models are promptly aligned with the most recent state of the global model after each local update, thereby maintaining stability and consistency during training. The second component is the adaptive weight function, which dynamically adjusts the distillation weights at different training stages to prevent client models from being misled by the global model’s poor performance in the early stages of training.

\begin{algorithm}[t]
\caption{FedADT}
\textbf{Input}:  all clients $c$,client learning rate $\eta$, client SGD steps $Q$, value calculation function $\beta$, global timestamp $t_g$ \\
\textbf{Output}: a optimized model
\begin{algorithmic}[1]
\STATE $t_g = 0$
\REPEAT
    \STATE $c \gets \text{sample clients}$ 
    \STATE \text{Run Client-Training}$(w^t, \eta, Q)$ on $c$ 
    \IF{\text{receive client update}}
        \STATE ($w_i,t_i) \gets \text{received from client } i$
        \STATE $\tau_i \gets t_g - t_i$
        \IF{$\tau > 1$}
            \STATE $w_i \gets Kowledge Distillation(w_g,w_i)$
        \ENDIF
        \STATE $\beta_i \gets \beta(\tau_i)$
        \STATE $w_g \gets (1 - \beta)w_g + \beta w_i$
        \STATE $t_g \gets t_g + 1$ 
    \ENDIF
\UNTIL{Convergence}
\end{algorithmic}
\end{algorithm}

\subsection{Knowledge Distillation}
In asynchronous federated learning, the models uploaded by different clients may not always correspond to the latest version of the global model. Such discrepancies in model versions can significantly impact the aggregation effectiveness and training performance of the global model. To address this issue, FedADT introduces a knowledge distillation mechanism. This mechanism utilizes the latest global model on the server to correct outdated models uploaded by clients, thereby reducing the interference caused by version differences during the training process and improving the overall system performance and convergence speed. For client-uploaded models that are not the latest version, we use the latest global model on the server to perform version correction through knowledge distillation, thereby reducing the impact of model version differences on the performance of asynchronous federated learning. Before training begins, the server needs to prepare a small labeled dataset for knowledge distillation of client models. This dataset can be extracted from the server's existing data or collected from other sources. To ensure the fairness and accuracy of the testing results, this distillation dataset must be completely independent of the test set. Additionally, the loss function used in our knowledge distillation process is defined as follows:
\begin{equation}
    \mathcal{L}_{\text{KD}} = \Tilde{\alpha} \cdot \text{KL} \left( \sigma \left( \frac{z_S}{T} \right) \middle\| \sigma \left( \frac{z_C}{T} \right) \right) + (1 - \Tilde{\alpha}) \cdot ( -\log \sigma(z_C)),
\end{equation}
where \( z_S \) and \( z_C \) represent the logits of the server-side model and the client-side model, respectively. \( T \) is the temperature parameter for knowledge distillation, which is used to smooth the probability distribution. $\sigma(\cdot)$ represents the Softmax function, defined as $\sigma(z_i) = \frac{\exp(z_i)}{\sum_j \exp(z_j)}$. $\text{KL}(P \| Q)$ represents the Kullback-Leibler divergence, which is used to measure the difference between two probability distributions $P$ and $Q$, defined as $\text{KL}(P \| Q) = \sum_i P(i) \log \frac{P(i)}{Q(i)}$. 

We recommend minimizing the size of the dataset used for knowledge distillation and the number of distillation rounds, as the global model continues to be updated during this process. It should be noted that in our experiments, we set the data ratio for knowledge distillation to 0.5\% and limited the distillation process to a single round. Since knowledge distillation itself achieves knowledge transfer at a relatively fast speed, we can ensure that, with a smaller amount of data, the time spent on distillation is significantly shorter than the time required for local training on the client side. This helps avoid the negative impact of significant changes in the global model version during the distillation process on training.

\subsection{Adaptive Weight Function}
Unlike traditional methods that rely on pre-trained teacher models, our knowledge distillation teacher model is a server-side model that is continuously updated during training. Therefore, using a fixed knowledge distillation weight $\alpha$ throughout the process is not appropriate. To better adapt to the dynamic nature of training, the weight of knowledge distillation should be reduced during the early stages of client training to mitigate potential misleading effects from the global model. In the mid-to-late stages of training, the weight should gradually return to a normal level to fully utilize the corrective effect of knowledge distillation on outdated client models. Our weighting function is designed as follows:
\begin{equation}
    \Tilde{\alpha}(t) = \Tilde{\alpha}_{min} + (\Tilde{\alpha}_{max} - \Tilde{\alpha}_{min}) \cdot \min\left(1, \frac{t}{T_g}\right).
\end{equation}

Among them, $\alpha_{\text{min}}$ and $\alpha_{\text{max}}$ represent the predefined minimum and maximum values of the knowledge distillation weight, respectively. $T_g$ denotes the preset number of training rounds for weight adjustment, and $t$ represents the current training round. Using this weighting function, we can dynamically adjust the weight during the first $T_g$ training rounds, thereby accelerating the convergence of the model.

\section{Experiment}
In this section, we conduct experiments using four different datasets and compare the proposed method with six other federated learning algorithms. The experimental results demonstrate that the proposed method outperforms other algorithms across various experimental settings, showing particularly significant advantages when compared with asynchronous federated learning algorithms. These findings indicate that our method has achieved a major breakthrough in the field of asynchronous federated learning.

\begin{table*}[t]
\centering
\caption{\textbf{Virtual time (seconds) required for different methods to achieve target accuracy.}}
\label{tab:timeresults}
\resizebox{\textwidth}{!}{%
\begin{tabular}{lccc|ccc|ccc|ccc}
\toprule
\multirow{3}{*}{\textbf{Methods}} & \multicolumn{3}{c|}{\textbf{MNIST}} & \multicolumn{3}{c|}{\textbf{FMNIST}} & \multicolumn{3}{c|}{\textbf{CIFAR10}} & \multicolumn{3}{c}{\textbf{CIFAR100}} \\ 
                 & \multicolumn{3}{c|}{\textit{Target Acc: 0.90}} & \multicolumn{3}{c|}{\textit{Target Acc: 0.75}} & \multicolumn{3}{c|}{\textit{Target Acc: 0.40}} & \multicolumn{3}{c}{\textit{Target Acc: 0.10}} \\
\cmidrule(r){2-13}
                 & $\alpha=0.1$            & $\alpha=0.5$       & $\alpha=1.0$             & $\alpha=0.1$           & $\alpha=0.5$             & $\alpha=1.0$           & $\alpha=0.1$             & $\alpha=0.5$           & $\alpha=1.0$             & $\alpha=0.1$           & $\alpha=0.5$  &$\alpha=1.0$ \\
\midrule
FedAVG           &397458           & 250312          &128982          & 104299        & 78213          &59630           &Fail      & 734126        & 521661           &Fail          & 980473          &918653              \\
FedAsync         &208658               & 95836           & 50883           &42780           & 35831           & 18938           & Fail            &Fail           &254897           &Fail           &Fail   &Fail                \\
FedBuff          &65059           &37194           &23102           & 27350           &15942           &10394           &349026          &193215           &148721          &515938           &322954     &294161               \\
CA2FL            & 62772           &34749           &25212          &Fail           &345237           & 179075           &1145905       &581034          & 211775         &Fail      &Fail  & 485448                \\
FedFa            &288541           & 193813           & 67087           &53401           & 394106           &28637           &Fail            &Fail           &553746           &Fail           &Fail   &Fail                     \\
FedAVGM           &397458           &229535          &104299           &124056           &74398          &59630           &Fail         &639136        &516679          &Fail  &728312 &893856          \\
\textbf{FedADT}             & \textbf{44136}  & \textbf{28493}  & \textbf{20905}  & \textbf{14391}  & \textbf{11324}  & \textbf{8269}  & \textbf{251159}  & \textbf{112764}  & \textbf{89373}  & \textbf{399009}   & \textbf{239431}  & \textbf{142346}   \\ 
\bottomrule
\end{tabular}%
}
\label{table:table1}
\end{table*}

\begin{table*}[t]
\centering
\caption{\textbf{Final accuracy of different algorithms under various settings after running for 15 days.}}
\label{tab:highresults}
\resizebox{\textwidth}{!}{%
\begin{tabular}{lccc|ccc|ccc|ccc}
\toprule
\multirow{3}{*}{\textbf{Methods}} & \multicolumn{3}{c|}{\textbf{MNIST}} & \multicolumn{3}{c|}{\textbf{FMNIST}} & \multicolumn{3}{c|}{\textbf{CIFAR10}} & \multicolumn{3}{c}{\textbf{CIFAR100}} \\ 
\cmidrule(r){2-13}
                 & $\alpha=0.1$            & $\alpha=0.5$       & $\alpha=1.0$             & $\alpha=0.1$           & $\alpha=0.5$             & $\alpha=1.0$           & $\alpha=0.1$             & $\alpha=0.5$           & $\alpha=1.0$             & $\alpha=0.1$           & $\alpha=0.5$  &$\alpha=1.0$        \\ \midrule
FedAVG           & 94.64           & 96.33           & 97.32           & 81.43           & 82.03          & 82.69           & 37.82           & 45.50          & 51.16           & 7.88           & 10.83    &12.24              \\
FedAsync           & 93.28           &95.38           & 96.69           & 81.58           & 81.70           & 82.30           & 28.85           & 39.84           & 43.88           & 4.41           & 7.53     & 8.33               \\
FedBuff          & 97.16           & 97.22           &97.65          &\textbf{83.18}          &\underline{82.98}          &\underline{82.72}           & \underline{42.21}           & \underline{48.36}          & 51.82          & \textbf{16.29}           & \textbf{21.78}                &\textbf{24.94}     \\
CA2FL           & \textbf{99.16}           & \textbf{99.32}           & \textbf{99.18}          & 74.35          & 75.61           &77.36         & 
36.30          & 47.33         &\textbf{55.89}         &6.44         &9.78       &14.87           \\
FedFa           & 92.41           &94.23           & 95.90           & 81.32           & 81.98           & 82.09           & 23.67           &33.18           &41.97           & 4.65           & 5.92         & 7.58            \\
FedAVGM           & 94.74           & 96.36           &97.31           &81.44         &82.09         & 82.68         & 38.14           &44.56           &51.08            & 7.85           &10.15      &12.24            \\
\textbf{FedADT}           & \underline{98.49}  & \underline{98.73}  &\underline{98.79}   & \underline{82.37}  & \textbf{83.09}  & \textbf{83.11}  & \textbf{48.55}  & \textbf{51.23}  & \underline{53.73}  &\underline{11.93}  &\underline{17.52}  & \underline{19.59}  \\ 
\bottomrule
\end{tabular}%
}
\label{table:table2}
\end{table*}

\begin{table}[t]
\centering
\caption{\textbf{Accuracy at round $T_g$.}}
\label{tab:finalresults}
\resizebox{\columnwidth}{!}{%
\begin{tabular}{lcc|cc|cc}
\toprule
\multirow{2}{*}{\textbf{Strategy}} & \multicolumn{2}{c|}{\textbf{MNIST}} & \multicolumn{2}{c|}{\textbf{FMNIST}} & \multicolumn{2}{c}{\textbf{CIFAR10}}\\ 
\cmidrule(r){2-7}
                & $\alpha=0.1$            & $\alpha=0.5$      & $\alpha=0.1$            & $\alpha=0.5$       & $\alpha=0.1$  & $\alpha=0.5$       \\ \midrule
Fixed $\Tilde{\alpha}$=0.2         & 62.18           & 65.92           & 72.21           & 72.27           & 14.32           & 16.24       \\
Fixed $\Tilde{\alpha}$=0.6          &84.59           &88.06           &76.74           &77.49           &23.29           &23.82                      \\
\textbf{Adaptive}          & \textbf{87.98}           & \textbf{91.04}           & \textbf{78.59}           & \textbf{79.33}           & \textbf{28.92}           & \textbf{30.46}                      \\
\bottomrule
\end{tabular}%
}
\label{table:table3}
\end{table}

\subsection{Experimental Setup}
We implement our algorithm FedADT using FLGO \cite{wang2023flgofullycustomizablefederated} and PyTorch. FLGO is a good federated learning framework that supports asynchronous federated learning. Our experimental setup is equipped with an NVIDIA GeForce RTX 3090 GPU.

\textbf{Settings.} We simulate a scenario with 500 clients. Our concurrency rate is set to 20\%. When the number of concurrent clients falls below this value, the server resamples 50 clients until the concurrency requirement is met, and each client's response time uniformly distributed between 0 and 5000. The entire training process lasted for 15 days of virtual time.

\textbf{Datasets and Models.} We conduct experiments on four representative datasets: MNIST \cite{lecun2010mnist}, FMNIST \cite{xiao2017/online}, CIFAR10 \cite{Krizhevsky09learningmultiple}, and CIFAR100 \cite{Krizhevsky09learningmultiple}. We extract 0.5\% of the data from the clients' datasets as knowledge distillation data, which was independent of the server's test set. We use CNN models for experiments on these datasets. For data partitioning, we use a Dirichlet distribution to allocate data among all clients, with $\alpha$ values set to $0.1$, $0.5$ and $1$, simulating both ideal and non-iid scenarios.

\textbf{Hyperparameters.} In all our experiments, the learning rate is set to $0.01$, with a decay rate of $0.9999$, and the batch size was $32$. For semi-asynchronous algorithms with a buffer, the buffer size is set to the optimal value of $10$ as suggested in the literature, and the remaining parameters are set to the recommended values from the original paper. The knowledge distillation rounds of our method are set to 1, $\beta$ is set to $\frac{1}{\sqrt{\tau + 1}}$, the distillation temperature $T$ is set to 3, $\Tilde{\alpha}_{\text{min}}$ is set to 0.2, $\Tilde{\alpha}_{\text{max}}$ is set to 0.6, and $T_g$ is set to 1000.

\textbf{Baselines.} To verify the performance of our method in asynchronous environments, we compare it with several classic and advanced federated learning algorithms across multiple datasets. The algorithms compared include FedAVG~\cite{mcmahan2023communicationefficientlearningdeepnetworks}, FedAVGM~\cite{hsu2019measuringeffectsnonidenticaldata}, FedAsync~\cite{xie2020asynchronousfederatedoptimization}, FedFa~\cite{10.24963/ijcai.2024/584}, FedBuff~\cite{nguyen2022federatedlearningbufferedasynchronous}, and CA2FL~\cite{wang2024tackling}. Among these, FedAVG and FedAVGM represent synchronous methods, which require waiting for all clients to complete their training before performing model aggregation. Although this introduces delays, their accuracy remains higher due to the absence of staleness issues. FedAsync and FedFa, as representatives of asynchronous methods, immediately aggregate once client updates are received. However, their accuracy is lower due to the impact of staleness. On the other hand, FedBuff and CA2FL, as semi-asynchronous methods, use buffering to balance speed and accuracy, achieving superior performance in environments with uneven delay distributions.

\subsection{Convergence Effectiveness}
We evaluate the convergence rate of FedADT by comparing it with different baseline methods in terms of the virtual time required to reach the target accuracy, with the specific results shown in Table~\ref{table:table1}. The primary goal of asynchronous algorithms is to minimize the time required to reach the target accuracy in order to leverage their speed advantage over synchronous methods, making convergence rate the most important evaluation metric for asynchronous approaches. Experimental results show that, under all experimental settings, our method consistently exhibits faster convergence than the other methods, highlighting its superiority.

\subsection{Performance Comparison}
\begin{figure}[t]
\hspace{-0.5cm}
    \centering
    \begin{minipage}{0.24\textwidth}
        \includegraphics[width=\linewidth, trim=0 0 0 0, clip]{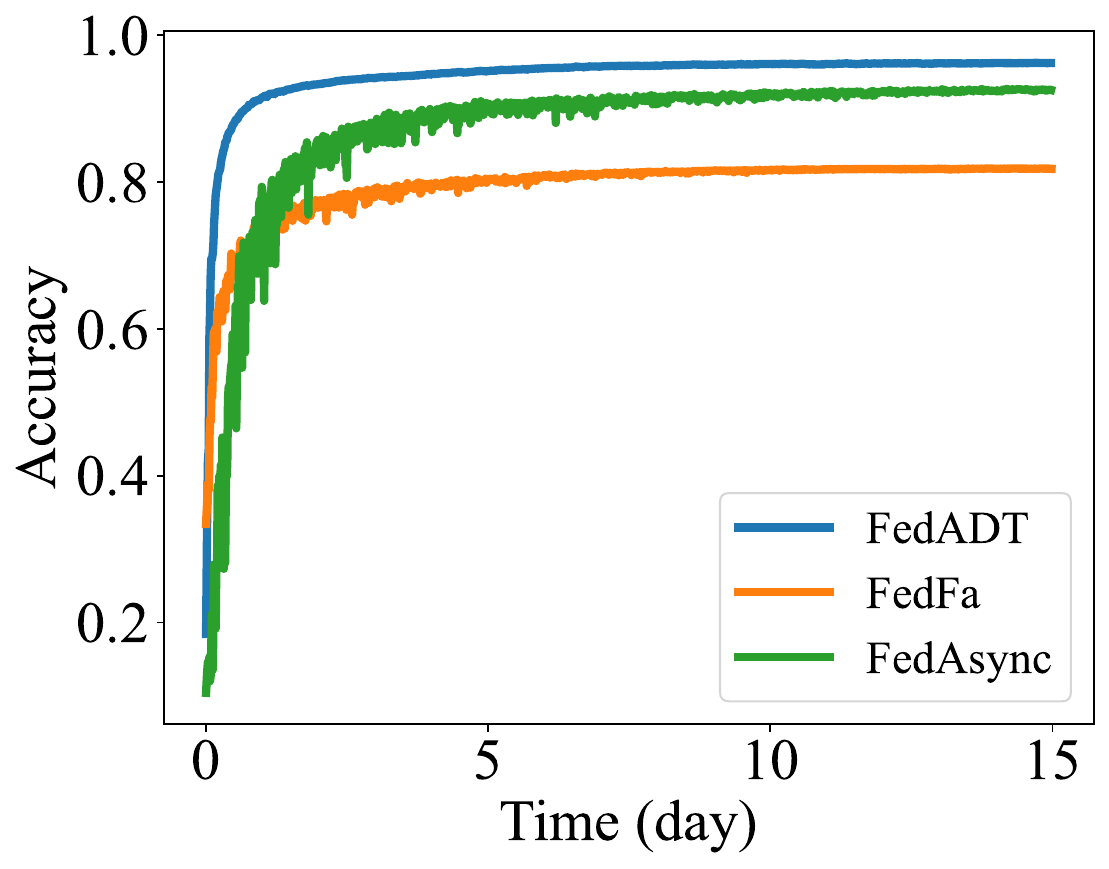}
        \subcaption{MNIST}  
    \end{minipage}%
    \begin{minipage}{0.24\textwidth}
        \includegraphics[width=\linewidth, trim=0 0 0 0, clip]{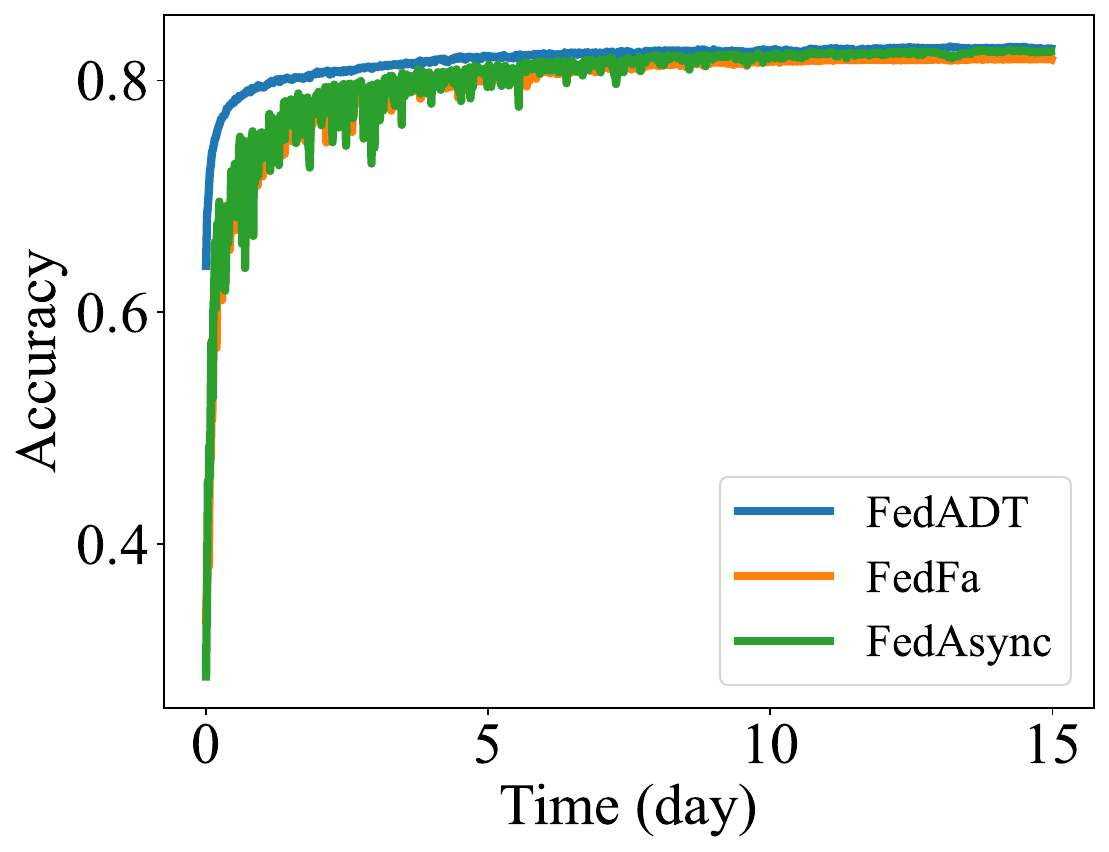}
        \subcaption{FMNIST}
    \end{minipage}%
    \\
\hspace{-0.5cm}
    \begin{minipage}{0.24\textwidth}
        \includegraphics[width=\linewidth, trim=0 0 0 0, clip]{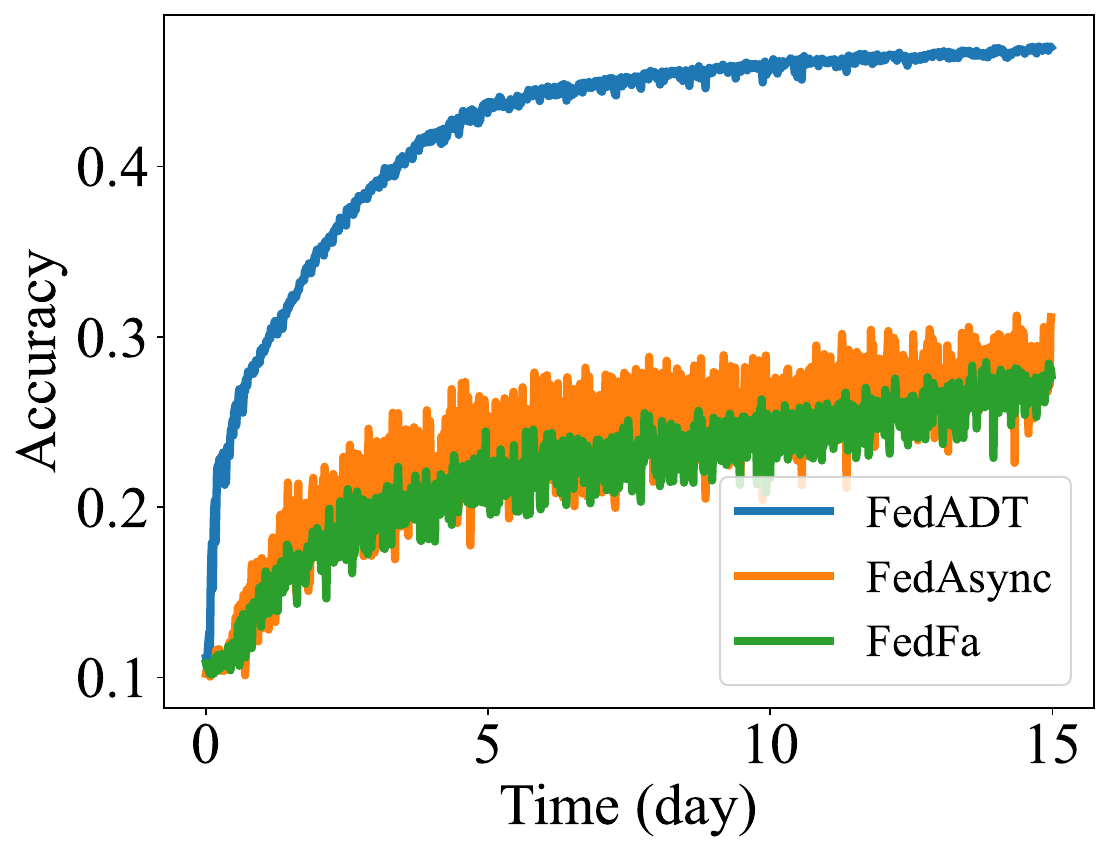}
        \subcaption{CIFAR10}  
    \end{minipage}%
    \begin{minipage}{0.24\textwidth}
        \includegraphics[width=\linewidth, trim=0 0 0 0, clip]{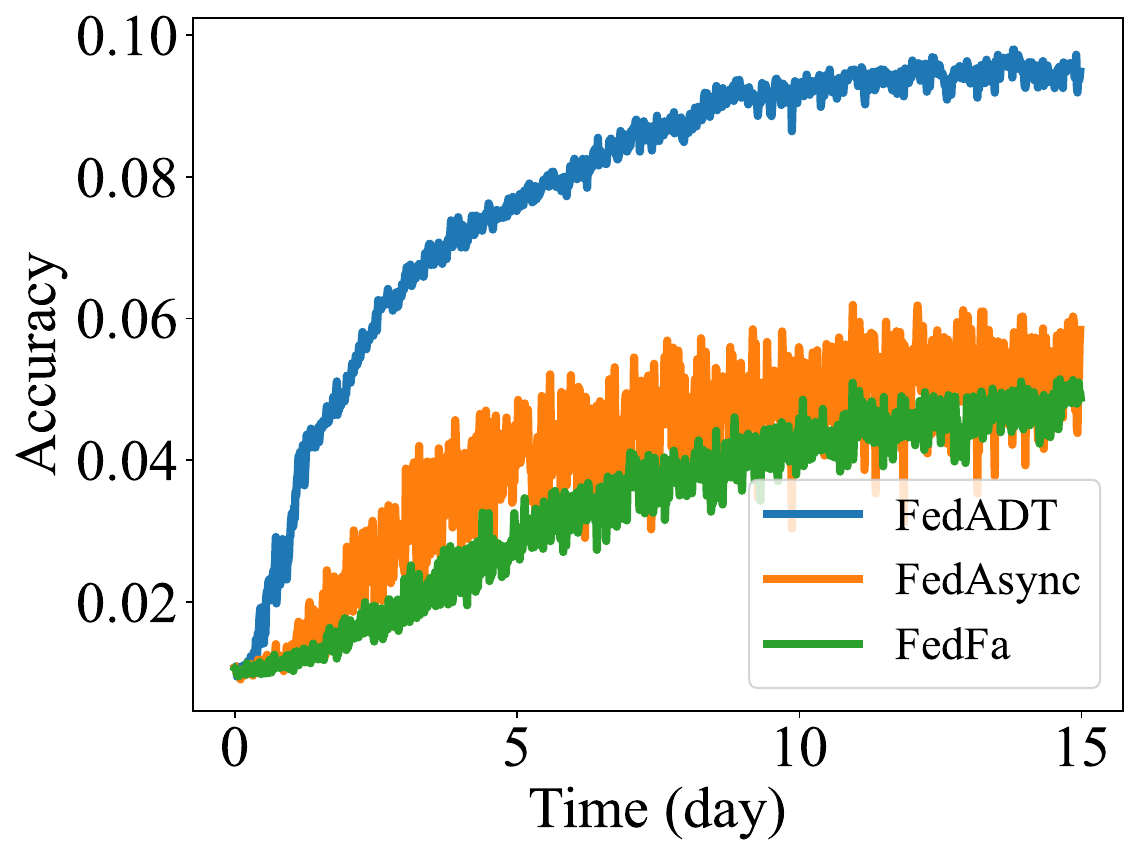}
        \subcaption{CIFAR100}
    \end{minipage}%
    \caption{
\textbf{Comparison with other pure asynchronous methods when $\alpha$ = 0.1.}}
    \label{fig:figure3}
    \vspace{-0.2cm}
\end{figure}

We conduct experiments on all baseline methods under the condition of fixed virtual runtime, and record the final accuracy achieved by each algorithm, as shown in Table~\ref{table:table2}. The experimental results demonstrate that our algorithm achieved optimal or near-optimal accuracy across all experimental settings. Due to the excellent buffer configuration of semi-asynchronous federated learning, our algorithm's accuracy is not optimal in some settings. However, our algorithm significantly outperforms other asynchronous algorithms, as shown in Fig.~\ref{fig:figure3}, and achieving performance comparable to or even surpassing that of semi-asynchronous federated learning algorithms. We believe that we have made a significant advancement in asynchronous federated learning, making robust asynchronous learning possible.

\subsection{Effectiveness of Adaptive Weight}
In this section, we investigate the impact of the adaptive weighting function on our algorithm. We test the model's accuracy at the $T_g$ round under three conditions: the knowledge distillation weight fixed at 0.2, fixed at 0.6, and using the adaptive weight function. The results are shown in Table~\ref{table:table3}. The experimental results demonstrate that our adaptive weighting function effectively facilitates a rapid warm-up of the model, thereby improving the convergence rate of the model.

\subsection{Impact of Concurrency Rate}
The concurrency rate in asynchronous methods is defined as the minimum proportion of concurrent clients to the total number of clients. In our algorithm, when the number of concurrent clients falls below the specified concurrency rate, the server resamples clients until the required number of concurrent clients is reached. The concurrency rate for asynchronous algorithms is typically set between 10\% and 20\%. This is because a high concurrency rate increases the upper limit of staleness, which reduces the final model accuracy, while a low concurrency rate results in prolonged server waiting times, negatively impacting the convergence speed. We compared the accuracy of our method with other asynchronous algorithms under different sampling rates $\mathcal{P}$, after 10 times the number of $T_g$ iterations. The results are shown in Table~\ref{table:table4}. Experimental results demonstrate that our algorithm exhibits greater robustness at higher concurrency rates, as knowledge distillation effectively mitigates the adverse effects of staleness on model training.

\begin{table}[t]
\centering
\caption{\textbf{Performance comparison under same rounds.}}
\begin{tabular}{lcccc}
\toprule
$\mathcal{P}$ & 10\% & 20\% & 25\% & 30\% \\
\midrule
FedAsync      & 0.6103 & 0.5935 & 0.5713 & 0.5587 \\
FedFa     &0.5984 & 0.5839 & 0.5640 & 0.5509 \\
\textbf{FedADT} & \textbf{0.6832}  & \textbf{0.6794} & \textbf{0.6713}  & \textbf{0.6642} \\
\bottomrule
\end{tabular}
\label{tab:performance}
\label{table:table4}
\end{table}

\section{Conclusion}
In this paper, we propose FedADT, a version correction method based on knowledge distillation, aimed at reducing the negative impact of version discrepancies in asynchronous federated learning. FedADT rapidly transfers the knowledge of the latest model to outdated client models through knowledge distillation, enabling the outdated models to be effectively guided toward the direction of the latest model, thereby reducing conflicts between different model versions. Additionally, by introducing an adaptive weight function, the performance of the method is further enhanced. Extensive experiments demonstrate that FedADT outperforms other asynchronous methods, with a faster convergence rate.

\section{Acknowledgement}
This work is supported by the Fundamental Research Funds for Central Universities under grant No. xzy012024105.  
\bibliographystyle{IEEEtran}
\bibliography{IEEEabrv,citation}
\end{document}